\documentclass[final]{cvpr}

\usepackage{times}
\usepackage{epsfig}
\usepackage{graphicx}
\usepackage{amsmath}
\usepackage{amssymb}
\usepackage{amsfonts}       
\usepackage{booktabs}       
\usepackage{multirow}
\usepackage{makecell}
\usepackage{algorithm,algorithmicx,algpseudocode}
\usepackage{pifont}

\usepackage[pagebackref=true,breaklinks=true,colorlinks,bookmarks=false]{hyperref}

\newcommand{\ts}[1]{{({#1})}}

\def\vphi{{\bm{\varphi}}}
\def\valpha{{\bm{\alpha}}}

\def\cvprPaperID{456} 
\def\confYear{CVPR 2021}

\usepackage{amsmath,amsfonts,bm}

\def\eqref#1{equation~\ref{#1}}

\def\1{\bm{1}}

\def\vzero{{\bm{0}}}

\def\vmu{{\bm{\mu}}}
\def\vtheta{{\bm{\theta}}}

\def\vw{{\bm{w}}}
\def\vx{{\bm{x}}}

\def\vz{{\bm{z}}}

\def\mI{{\bm{I}}}

\def\mX{{\bm{X}}}

\def\mSigma{{\bm{\Sigma}}}

\DeclareMathAlphabet{\mathsfit}{\encodingdefault}{\sfdefault}{m}{sl}
\SetMathAlphabet{\mathsfit}{bold}{\encodingdefault}{\sfdefault}{bx}{n}

\newcommand{\E}{\mathbb{E}}

\newcommand{\KL}{D_{\mathrm{KL}}}

\begin{document}

\title{Diffusion Probabilistic Models for 3D Point Cloud Generation}

\author{Shitong Luo, Wei Hu \thanks{Corresponding author: Wei Hu (forhuwei@pku.edu.cn). This work was supported by the National Key R\&D project of China under contract No. 2019YFF0302903 and National Natural Science Foundation of China under contract No. 61972009.}\\
Wangxuan Institute of Computer Technology\\
Peking University\\
{\tt\small \{luost, forhuwei\}@pku.edu.cn}
}

\maketitle

\thispagestyle{empty}

\begin{abstract}
We present a probabilistic model for point cloud generation, which is fundamental for various 3D vision tasks such as shape completion, upsampling, synthesis and data augmentation.  
Inspired by the diffusion process in non-equilibrium thermodynamics, we view points in point clouds as particles in a thermodynamic system in contact with a heat bath, which diffuse from the original distribution to a noise distribution. 
Point cloud generation thus amounts to learning the reverse diffusion process that transforms the noise distribution to the distribution of a desired shape. 
Specifically, we propose to model the reverse diffusion process for point clouds as a Markov chain conditioned on certain shape latent. We derive the variational bound in closed form for training and provide implementations of the model.
Experimental results demonstrate that our model achieves competitive performance in point cloud generation and auto-encoding. 
The code is available at \url{https://github.com/luost26/diffusion-point-cloud}.
\end{abstract}

\section{Introduction}
\label{sec:intro}
With recent advances in depth sensing and laser scanning, point clouds have attracted increasing attention as a popular representation for modeling 3D shapes. 
Significant progress has been made in developing methods for point cloud analysis, such as classification and segmentation \cite{qi2017pointnet, qi2017pointnet++, wang2019dgcnn}.
On the other hand, learning generative models for point clouds is powerful in unsupervised representation learning to characterize the data distribution, which lays the foundation for various tasks such as shape completion, upsampling, synthesis, \etc.  

Generative models such as variational auto-encoders (VAEs), generative adversarial networks (GANs), normalizing flows, \etc, have shown great success in image generation \cite{kingma2013vae, goodfellow2014gan, chen2016variational, dinh2016density}.    
However, these powerful tools cannot be directly generalized to point clouds, due to the irregular sampling patterns of points in the 3D space in contrast to regular grid structures underlying images. 
Hence, learning generative models for point clouds is quite challenging. 
Prior research has explored point cloud generation via GANs \cite{achlioptas2018rgan, valsesia2018gcngan, shu2019treegan}, auto-regressive models \cite{sun2020pointgrow}, flow-based models \cite{yang2019pointflow} and so on. 
While remarkable progress has been made, these methods have some inherent limitations for modeling point clouds. 
For instance, the training procedure could be unstable for GANs due to the adversarial losses.
Auto-regressive models assume a generation ordering which is unnatural and might restrict the model's flexibility.

\begin{figure}[]
\begin{center}
\includegraphics[width=0.45\textwidth]{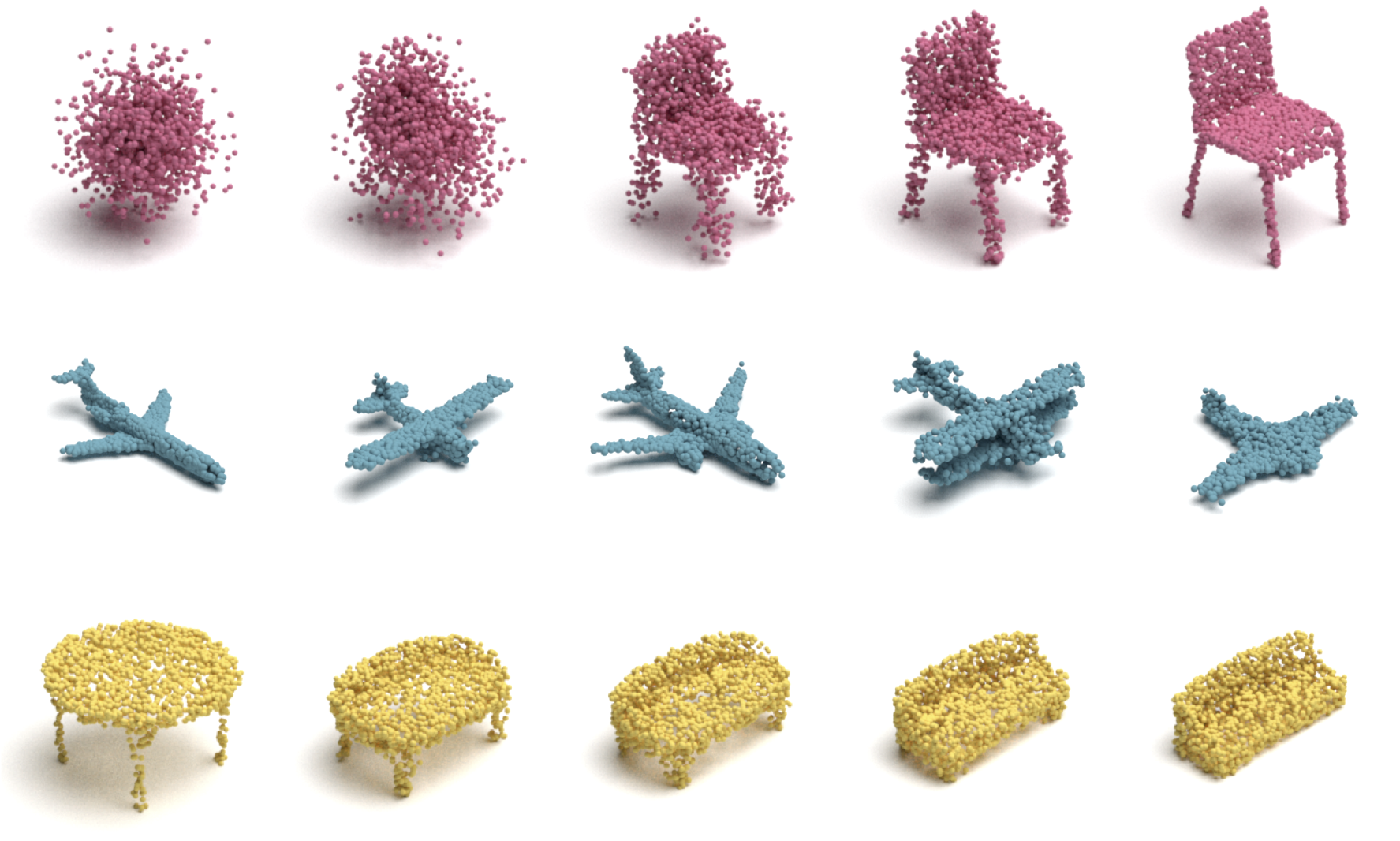}
\end{center}
\caption{\textbf{Top}: The diffusion process that converts noise to some shape (left to right). \textbf{Middle}: Generated point clouds from the proposed model. \textbf{Bottom}: Latent space interpolation between the two point clouds at both ends.}
\label{fig:teaser}
\end{figure}

In this paper, we propose a probabilistic generative model for point clouds inspired by non-equilibrium thermodynamics, exploiting the \textit{reverse diffusion process} to learn the point distribution. 
As a point cloud is composed of discrete points in the 3D space, we regard these points as particles in a non-equilibrium thermodynamic system in contact with a heat bath. 
Under the effect of the heat bath, the position of particles evolves stochastically in the way that they diffuse and eventually spread over the space. This process is termed the \textit{diffusion process} that converts the initial distribution of the particles to a simple noise distribution by adding noise at each time step \cite{jarzynski1997equilibrium, sohl2015deep}.
Analogously, we connect the point distribution of point clouds to a noise distribution via the diffusion process. 
Naturally, in order to model the point distribution for point cloud generation, we consider \textit{the reverse diffusion process},  which recovers the target point distribution from the noise distribution. 

In particular, we model this reverse diffusion process as a Markov chain that converts the noise distribution into the target distribution. Our goal is to learn its transition kernel such that the Markov chain can reconstruct the desired shape. 
Further, as the purpose of the Markov chain is modeling the point distribution, the Markov chain alone is incapable to generate point clouds of various shapes. 
To this end, we introduce a {\it shape latent} as the condition for the transition kernel. 
In the setting of generation, the shape latent follows a prior distribution which we parameterize via normalizing flows \cite{chen2016variational, dinh2016density} for strong model expressiveness. In the setting of auto-encoding, the shape latent is learned end-to-end.
Finally, we formulate the training objective as maximizing the variational lower bound of the likelihood of the point cloud conditional on the shape latent, which is further formulated into tractable expressions in closed form. 
We apply our model to point cloud generation, auto-encoding and unsupervised representation learning, and results demonstrate that our model achieves competitive performance on point cloud generation and auto-encoding and comparable results on unsupervised representation learning. 

Our main contributions include:
\begin{itemize}
\setlength{\itemsep}{0pt}
\setlength{\parskip}{0pt}
    \item We propose a novel probabilistic generative model for point clouds, inspired by the diffusion process in non-equilibrium thermodynamics.
    \item We derive a tractable training objective from the variational lower bound of the likelihood of point clouds conditioned on some shape latent.
    \item Extensive experiments show that our model achieves competitive performance in point cloud generation and auto-encoding. 
\end{itemize}

\section{Related Works}
\label{sec:related}

\paragraph{Point Cloud Generation}
Early point cloud generation methods \cite{achlioptas2018rgan, gadelha2018multiresolution} treat point clouds as $N \times 3$ matrices, where $N$ is the fixed number of points, converting the point cloud generation problem to a matrix generation problem, so that existing generative models are readily applicable. For example, \cite{gadelha2018multiresolution} apply variational auto-encoders \cite{kingma2013vae} to point cloud generation. \cite{achlioptas2018rgan} employ generative adversarial networks \cite{goodfellow2014gan} based on a pre-trained auto-encoder. The main defect of these methods is that they are restricted to generating point clouds with a fixed number of points, and lack the property of permutation invariance. 
FoldingNet and AtlasNet \cite{yang2018foldingnet, groueix2018atlasnet} mitigate this issue by learning a mapping from 2D patches to the 3D space, which deforms the 2D patches into the shape of point clouds. These two methods allow generating arbitrary number of points by first sampling some points on the patches and then applying the mapping on them. In addition, the points on the patches are inherently invariant to permutation.

The above methods rely on heuristic set distances such as the Chamfer distance (CD) and the Earth Mover's distance (EMD). As pointed out in \cite{yang2019pointflow}, CD has been shown to incorrectly favor point clouds that are overly concentrated in the mode of the marginal point distribution, and EMD is slow to compute while approximations could lead to biased gradients.

Alternatively, point clouds can be regarded as samples from a point distribution. This viewpoint inspires exploration on applying likelihood-based methods to point cloud modeling and generation. 
PointFlow \cite{yang2019pointflow} employs continous normalizing flows \cite{chen2018neuralode, grathwohl2018ffjord} to model the distribution of points. DPF-Net \cite{klokov2020discrete} uses affine coupling layers as the normalizing flow to model the distribution.
PointGrow \cite{sun2020pointgrow} is an auto-regressive model with exact likelihoods. 
More recently, \cite{cai2020shapegf} proposed a score-matching energy-based model ShapeGF to model the distribution of points.

Our method also regards point clouds as samples from a distribution, but differs in the probabilistic model compared to prior works. We leverage the reverse diffusion Markov chain to model the distribution of points, achieving both simplicity and flexibility. Specifically, the training process of our model involves learning the Markov transition kernel, whose training objective has a simple function form. By contrast, GAN-based models involve complex adversarial losses, continuous-flow-based methods involve expensive ODE integration. In addition, our model is flexible, because it does not require invertibility in contrast to flow-based models, and does not assume ordering compared to auto-regressive models.

\paragraph{Diffusion Probabilistic Models}
The diffusion process considered in this work is related to the diffusion probabilistic model \cite{sohl2015deep, ho2020denoising}. Diffusion probabilistic models are a class of latent variable models, which also use a Markov chain to convert the noise distribution to the data distribution. Prior research on diffusion probabilistic models focuses on the unconditional generation problem for toy data and images. 
In this work, we focus on point cloud generation, which is a {\it conditional} generation problem, because the Markov chain considered in our work generates points of a point cloud conditioned on some shape latent. This conditional adaptation leads to significantly different training and sampling schemes compared to prior research on diffusion probabilistic models.

\section{Diffusion Probabilistic Models for Point Clouds}
\label{sec:method}

\begin{figure*}[]
\begin{center}
\includegraphics[width=0.7\textwidth]{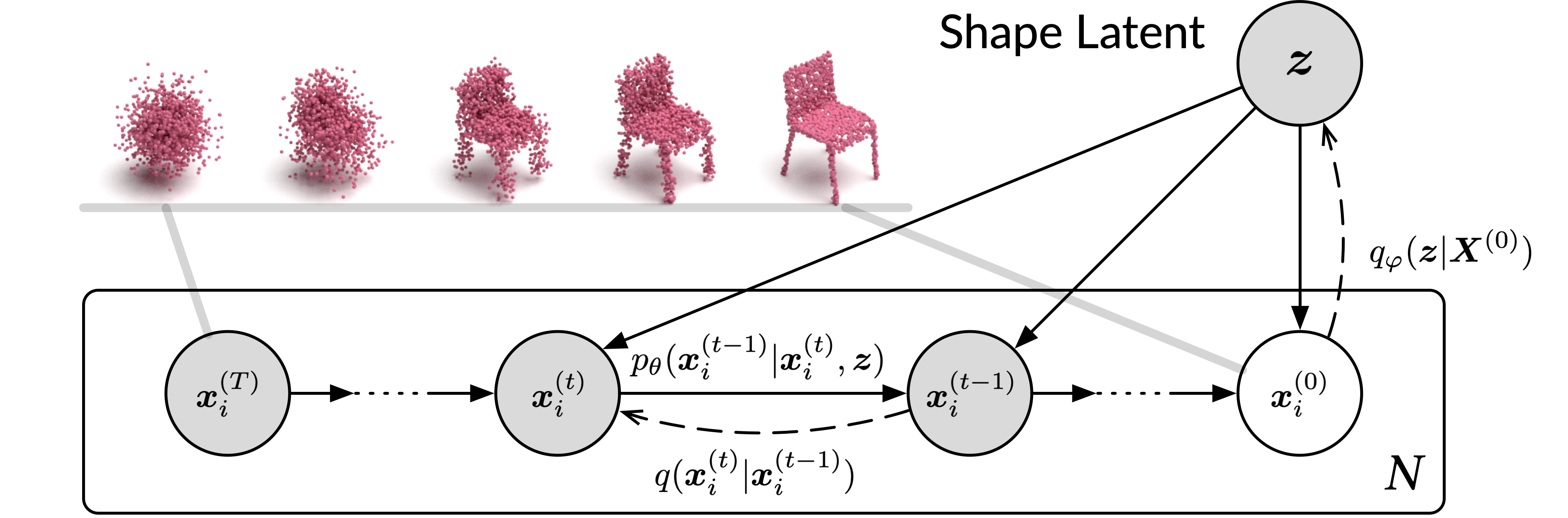}
\end{center}
\caption{The directed graphical model of the diffusion process for point clouds. $N$ is the number of points in the point cloud $\mX^\ts{0}$.}
\end{figure*}

In this section, we first formulate the probabilistic model of both the forward and the reverse diffusion processes for point clouds. 
Then, we formulate the objective for training the model.
The implementation of the model will be provided in the next section.

\subsection{Formulation}
\label{sec:diffusion}

We regard a point cloud $\mX^\ts{0} = \{ \vx_i^\ts{0} \}_{i=1}^{N}$ consisting of $N$ points as a set of particles in an evolving thermodynamic system. 
As discussed in Section~\ref{sec:intro}, each point $\vx_i$ in the point cloud can be treated as being sampled independently from a point distribution, which we denote as $q (\vx_i^{(0)} | \vz)$. Here, $\vz$ is the shape latent that determines the distribution of points. 

Physically, as time goes by, the points gradually diffuse into a chaotic set of points. 
This process is termed the {\it diffusion process}, which converts the original meaningful point distribution into a noise distribution. The forward diffusion process is modeled as a Markov chain \cite{jarzynski1997equilibrium}:
\begin{equation}
\label{eq:forward}
q(\vx_i^{(1 : T)} | \vx_i^{(0)})  = \prod_{t = 1}^{T} q(\vx_i^{(t)} | \vx_i^{(t-1)}) , 
\end{equation}
where $q(\vx_i^{(t)} | \vx_i^{(t-1)})$ is the Markov diffusion kernel. The kernel adds noise to points at the previous time step and models the distribution of points at the next time step.
Following the convention of \cite{sohl2015deep}, we define the diffusion kernel as:
\begin{equation}
\label{eq:forwardkrnl}
q(\vx^{(t)} | \vx^{(t-1)}) = \mathcal{N} (\vx^{(t)} | \sqrt{1-\beta_t} \vx^{(t-1)}, \beta_t \mI), t=1,...,T,
\end{equation}
where $\beta_1 \ldots \beta_T$ are variance schedule hyper-parameters that control the diffusion rate of the process.

Our goal is to generate point clouds with a meaningful shape, encoded by the latent representation $\vz$.
We treat the generation process as the reverse of the diffusion process, where points sampled from a simple noise distribution $p(\vx_i^\ts{T})$ that approximates $q(\vx_i^\ts{T})$ are given as the input. Then, the points are passed through the reverse Markov chain and finally form the desired shape. Unlike the forward diffusion process that simply adds noise to the points, the reverse process aims to recover the desired shape from the input noise, which requires training from data.
We formulate the reverse diffusion process for generation as:
\begin{equation}
\label{eq:reverse}
p_\vtheta(\vx^\ts{0 : T} | \vz) = p(\vx^\ts{T}) \prod_{t=1}^{T} p_\vtheta (\vx^\ts{t-1} | \vx^\ts{t} , \vz),
\end{equation}
\begin{equation}
\label{eq:revkrnl}
p_\vtheta (\vx^\ts{t-1} | \vx^\ts{t} , \vz) = \mathcal{N} \big( \vx^\ts{t-1} \big| \vmu_\vtheta(\vx^\ts{t}, t, \vz), \beta_t\mI  \big),
\end{equation}
where $\vmu_\vtheta$ is the estimated mean implemented by a neural network parameterized by $\vtheta$. 
$\vz$ is the latent encoding the target shape of the point cloud. 
The starting distribution $p(\vx_i^\ts{T})$ is set to a standard normal distribution $\mathcal{N}(\vzero, \mI)$. Given a shape latent $\vz$, we obtain the point cloud with the target shape by passing a set of points sampled from $p(\vx_i^\ts{T})$ through the reverse Markov chain.

For the sake of brevity, in the following sections, we use the distribution with respect to the entire point cloud $\mX^\ts{0}$. 
Since the points in a point cloud are independently sampled from a distribution, the probability of the whole point cloud is simply the product of the probability of each point:
\begin{equation}
\label{eq:independent}
    q(\mX^\ts{1:T} | \mX^{0}) = \prod_{i=1}^{N} q(\vx_i^{(1 : T)} | \vx_i^{(0)}),
\end{equation}
\begin{equation}
    p_\vtheta(\mX^\ts{0 : T} | \vz) = \prod_{i=1}^{N} p_\vtheta(\vx_i^\ts{0 : T} | \vz).
\end{equation}

Having formulated the forward and reverse diffusion processes for point clouds, we will formalize the training objective of the reverse diffusion process for point cloud generation as follows.

\subsection{Training Objective}
\label{sec:objective}

The goal of training the reverse diffusion process is to maximize the log-likelihood of the point cloud: $\E [ \log p_\vtheta (\mX^\ts{0}) ]$. 
However, since directly optimizing the exact log-likelihood is intractable, we instead \textit{maximize} its variational lower bound:
\begin{equation} 
\label{eq:elbo}
\begin{split}
    \E \big[ \log p_\vtheta (\mX^\ts{0}) \big] & \ge \E_q \Big[ \log \frac{p_\vtheta (\mX^\ts{0:T}, \vz)}{q(\mX^\ts{1:T}, \vz | \mX^\ts{0})} \Big] \\
    & = \E_q \Big[ \log p(\mX^\ts{T}) \\
    & \qquad + \sum_{t=1}^{T} \log \frac{p_\vtheta(\mX^\ts{t-1} | \mX^\ts{t}, \vz)}{q(\mX^\ts{t} | \mX^\ts{t-1})} \\
    & \qquad - \log \frac{q_\vphi(\vz | \mX^\ts{0})}{p(\vz)} \Big] .
\end{split}
\end{equation}
The above variational bound can be adapted into the training objective $L$ to be \textit{minimized} (the detailed derivation is provided in the supplementary material):
\begin{equation}
\label{eq:obj}
\begin{split}
    L(\vtheta, \vphi) & = \E_q \Big[ \sum_{t=2}^{T} \KL\big(q(\mX^\ts{t-1} | \mX^\ts{t}, \mX^\ts{0}) \| \\
    & \qquad \qquad \qquad \qquad p_\vtheta(\mX^\ts{t-1} | \mX^\ts{t}, \vz) \big) \\
    & \qquad - \log p_\vtheta(\mX^\ts{0} | \mX^\ts{1}, \vz) \\
    & \qquad + \KL\big( q_\vphi(\vz | \mX^\ts{0}) \| p(\vz) \big) \Big].
\end{split}
\end{equation}

Since the distributions of points are independent of each other as described in Eq.~(\ref{eq:independent}), we further expand the training objective:
\begin{equation}
\label{eq:objective_final}
\begin{split}
    L(\vtheta, \vphi) & = \E_q \Big[ \sum_{t=2}^{T} \sum_{i=1}^{N} \KL\big(
        \underbrace{q(\vx_i^\ts{t-1} | \vx_i^\ts{t}, \vx_i^\ts{0})}_{\textcircled{1}}  \| \\
    & \qquad \qquad \qquad \qquad 
        \underbrace{p_\vtheta(\vx_i^\ts{t-1} | \vx_i^\ts{t}, \vz) \big)}_{\textcircled{2}} \\
    & \qquad - \sum_{i=1}^{N} \underbrace{\log p_\vtheta(\vx_i^\ts{0} | \vx_i^\ts{1}, \vz)}_{\textcircled{3}} \\
    & \qquad + \KL\big( \underbrace{q_\vphi(\vz | \mX^\ts{0})}_{\textcircled{4}} \| \underbrace{p(\vz)}_{\textcircled{5}} \big)
    \Big].
\end{split}
\end{equation}
The training objective can be optimized efficiently since each of the terms on the right hand side is tractable and $q$ is easy to sample from the forward diffusion process.
Next, we elaborate on the terms to reveal how to compute the objective.

\textbf{\textcircled{1}} $q(\vx_i^\ts{t-1} | \vx_i^\ts{t}, \vx_i^\ts{0})$ can be computed with the following closed-form Gaussian according to \cite{ho2020denoising}:
\begin{equation}
\begin{split}
    q(\vx_i^\ts{t-1} | \vx_i^\ts{t}, \vx_i^\ts{0}) = \mathcal{N}(\vx_{i}^\ts{t-1} | \vmu_t(\vx^\ts{t}, \vx^\ts{0}), \gamma_t\mI),
\end{split}
\end{equation}
where, using the notation $\alpha_t = 1-\beta_t$ and $\bar{\alpha}_t = \prod_{s=1}^{t} \alpha_s$:
\begin{equation}
\label{eq:qrev}
\begin{split}
    \vmu_t (\vx^\ts{t}, \vx^\ts{0}) & = \frac{\sqrt{\bar{\alpha}_{t-1}} \beta_{t}}{1-\bar{\alpha}_{t}} \vx^\ts{0} + \frac{\sqrt{\alpha_{t}}\left(1-\bar{\alpha}_{t-1}\right)}{1-\bar{\alpha}_{t}} \vx^\ts{t}, \\
    \gamma_t & = \frac{1-\bar{\alpha}_{t-1}}{1-\bar{\alpha}_{t}} \beta_{t}.
\end{split}
\end{equation}

\textbf{\textcircled{2}, \textcircled{3}} $p_\vtheta(\vx_i^\ts{t-1} | \vx_i^\ts{t}, \vz) (t = 1, \ldots, T)$ are trainable Gaussians as described in Eq.~(\ref{eq:revkrnl}). 

\textbf{\textcircled{4}} $q_\vphi(\vz | \mX^\ts{0})$ is the approximate posterior distribution. 
Using the language of variational auto-encoders, $q_\vphi(\vz | \mX^\ts{0})$ is an encoder that encodes the input point cloud $\mX^\ts{0}$ into the distribution of the latent code $\vz$. 
We assume it as a Gaussian following the convention:
\begin{equation}
    q(\vz | \mX^\ts{0}) = \mathcal{N}\big(\vz | \vmu_\vphi(\mX^\ts{0}), \mSigma_\vphi(\mX^\ts{0}) \big).
\end{equation}

\textbf{\textcircled{5}} The last term $p(\vz)$ is the prior distribution. The most common choice of $p(\vz)$ is the isotropic Gaussian $\mathcal{N}(\vzero, \mI)$. In addition to a fixed distribution, the prior can be a trainable parametric distribution, which is more flexible. For example, normalizing flows \cite{chen2016variational, dinh2016density} can be employed to parameterize the prior distribution.

In the following section, we show how to optimize the objective in Eq.~(\ref{eq:objective_final}) in order to train the model.

\begin{figure*}[]
\begin{center}
\includegraphics[width=0.9\textwidth]{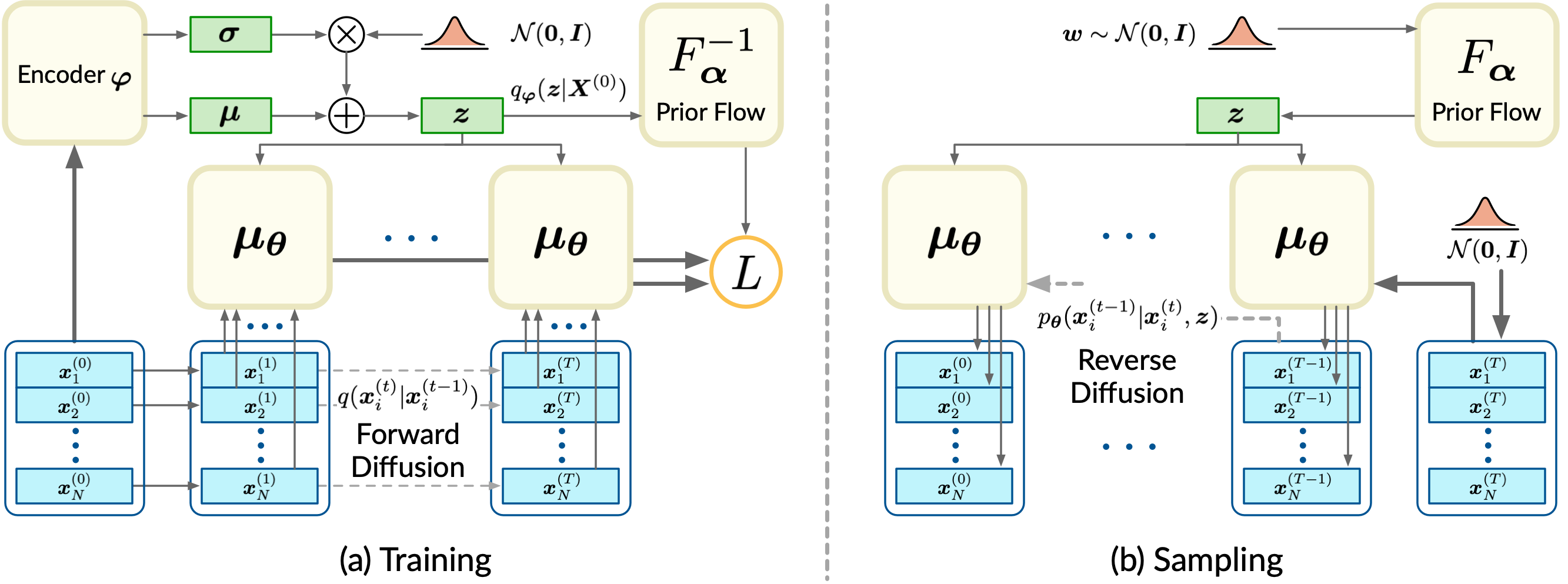}
\end{center}
\caption{The illustration of the proposed model. (a) illustrates how the objective is computed during the training process. (b) illustrates the generation process.}
\end{figure*}

\subsection{Training Algorithm}

In principle, training the model amounts to minimizing the objective in Eq.~(\ref{eq:objective_final}).
However, evaluating Eq.~(\ref{eq:objective_final}) requires summing the expectation of the KL terms over all the time steps, which involves sampling a full trajectory $\vx_i^\ts{1}, \ldots, \vx_i^\ts{T}$ from the forward diffusion process in order to compute the expectation.

To make the training simpler and more efficient, following \cite{ho2020denoising}, instead of evaluating the expectation of the whole summation over all the time steps in Eq.~(\ref{eq:objective_final}), we randomly choose one term from the summation to optimize at each training step.

Specifically, this simplified training algorithm is as follows:
\algrenewcommand\algorithmicindent{1.0em}%
\begin{algorithm}[H]
  \caption{Training (Simplified)} \label{alg:trainingmain}
  \small
  \begin{algorithmic}[1]
    \Repeat
      \State Sample $\mX^\ts{0} \sim q_\mathrm{data}(\mX^\ts{0})$
      \State Sample $\vz \sim q_\vphi(\vz | \mX^\ts{0})$
      \State Sample $t \sim \operatorname{Uniform}(\{ 1, \ldots, T\})$
      \State Sample $\vx_1^\ts{t},\ldots,\vx_N^\ts{t} \sim q(\vx^\ts{t} | \vx^\ts{0})$ 
      \State $L_t \gets \sum_{i=1}^{N} \KL\left( q(\vx_i^\ts{t-1} | \vx_i^\ts{t}, \vx_i^\ts{0}) \Big\| p_\vtheta(\vx_i^\ts{t-1} | \vx_i^\ts{t}, \vz) \big) \right)$
      \State $L_{\vz} \gets \KL( q_\vphi(\vz | \mX^\ts{0}) \| p(\vz)) $
      \State Compute $\nabla_\vtheta (L_t + \frac{1}{T} L_{\vz})$. Then perform gradient descent.
    \Until{converged}
  \end{algorithmic}
\end{algorithm}

To efficiently sample from $q(\vx^\ts{t} | \vx^\ts{0})$ (5th statement) and avoid iterative sampling starting from $t=1$, we leverage on the result in \cite{ho2020denoising}, which shows that $q(\vx^\ts{t} | \vx^\ts{0})$ is a Gaussian:
\begin{equation}
    q(\vx^\ts{t} | \vx^\ts{0}) = \mathcal{N} ( \vx^\ts{t} | \sqrt{\bar{\alpha}_t} \vx^\ts{0}, (1 - \bar{\alpha}_t) \mI ). 
\end{equation}
The gaussianity of $q(\vx^\ts{t} | \vx^\ts{0})$ makes further simplification on $L_t$ (6th statement) possible by using the reparameterization trick \cite{kingma2013vae}. 
We put the detail of this simplification to the supplementary material. Last, note that the KL divergence in $L_\vz$ is evaluated stochastically by $- \E_{\vz \sim q_\vphi(\vz | \mX^\ts{0})}\big[ p(\vz) \big] - H\big[q_\vphi(\vz | \mX^\ts{0})\big]$.

\section{Model Implementations}
\label{sec:impl}

The general training objective and algorithm in the previous section lay the foundation for the formulation of specific point cloud tasks. 
Next, we adapt the training objective to point cloud generation and point cloud auto-encoding respectively. 

\subsection{Point Cloud Generator}

Leveraging on the model in Section~\ref{sec:method}, we propose a probabilistic model for point cloud generation by employing normalizing flows to parameterize the prior distribution $p(\vz)$, which makes the model more flexible \cite{rezende2015variational, chen2016variational}.

Specifically, we use a stack of affine coupling layers \cite{dinh2016density} as the normalizing flow. In essence, the affine coupling layers provide a trainable bijector $F_\valpha$ that maps an isotropic Gaussian to a complex distribution. Since the mapping is bijective, the exact probability of the target distribution can be computed by the change-of-variable formula:
\begin{equation}
\label{eq:priorflow}
    p(\vz) = p_\vw(\vw) \cdot \bigg| \det \frac{\partial F_\valpha}{\partial \vw} \bigg|^{-1} \quad \text{where} \quad \vw = F_\valpha^{-1}(\vz).
\end{equation}
Here, $p(\vz)$ is the prior distribution in the model, $F_\valpha$ is the trainable bijector implemented by the affine coupling layers, and $ p_\vw(\vw) $ is the isotropic Gaussian $ \mathcal{N}(\vzero, \mI)$.

As for the encoder $q_\vphi(\vz | \mX^\ts{0})$, we adopt PointNet \cite{qi2017pointnet} as the architecture for $\vmu_\vphi$ and $\mSigma_\vphi$ of the encoder $q_\vphi(\vz | \mX^\ts{0})$.

Substituting Eq.~(\ref{eq:priorflow}) into Eq.~(\ref{eq:objective_final}), the training objective for the generative model is:
\begin{equation}
\label{eq:objgen}
\begin{split}
    L_G(\vtheta, \vphi, \valpha) & = \E_q \Big[ \sum_{t=2}^{T} \sum_{i=1}^{N} \KL\big(q(\vx_i^\ts{t-1} | \vx_i^\ts{t}, \vx_i^\ts{0}) \| \\
    & \qquad \qquad \qquad \qquad p_\vtheta(\vx_i^\ts{t-1} | \vx_i^\ts{t}, \vz) \big) \\
    & \quad - \sum_{i=1}^{N} \log p_\vtheta(\vx_i^\ts{0} | \vx_i^\ts{1}, \vz) \\ 
    & \quad + \KL\Big( q_\vphi(\vz | \mX^\ts{0}) \Big\| p_\vw(\vw) \cdot \Big| \det \frac{\partial F_\valpha}{\partial \vw} \Big|^{-1} \Big)
    \Big].
\end{split}
\end{equation}
The algorithm for optimizing the above objective can be naturally derived from Algorithm~\ref{alg:trainingmain}. 

To sample a point cloud, we first draw $\vw \sim \mathcal{N}(\vzero, \mI)$ and pass it through $F_\valpha$ to acquire the shape latent $\vz$. Then, with the shape latent $\vz$, we sample some points $\{ \vx_i^\ts{T}\}$ from the noise distribution $p(\vx^\ts{T})$ and pass the points through the reverse Markov chain $p_\vtheta(\vx_i^\ts{0:T} | \vz)$ defined in Eq.~(\ref{eq:reverse}) to generate the point cloud $\mX^\ts{0}=\{ \vx_i^\ts{0} \}_{i=1}^{N} $.

\begin{table*}[t]
\caption{Comparison of point cloud generation performance. CD is multiplied by $10^3$, EMD is multiplied by $10^1$, and JSD is multiplied by $10^3$.}
\label{table:generate}
\begin{center}
\begin{tabular}{ll|cc|cc|cc|c}
\toprule
 &  & \multicolumn{2}{c|}{MMD ($\downarrow$)} & \multicolumn{2}{c|}{COV (\%, $\uparrow$)} & \multicolumn{2}{c|}{1-NNA (\%, $\downarrow$)} & JSD ($\downarrow$) \\ \cmidrule{3-9}
Shape & Model & CD & EMD & CD & EMD & CD & EMD & - \\
\midrule

\multirow{7}{*}{Airplane} 
 & PC-GAN \cite{achlioptas2018rgan}
    & 3.819 & 1.810 & 42.17 & 13.84 & 77.59 & 98.52 &  6.188  \\
 & GCN-GAN \cite{valsesia2018gcngan}
    & 4.713 & 1.650 & 39.04 & 18.62 & 89.13 & 98.60 &  6.669 \\
 & TreeGAN \cite{shu2019treegan}
    & 4.323 & 1.953 & 39.37 &  8.40 & 83.86 & 99.67 &  15.646\\
 & PointFlow \cite{yang2019pointflow}
    & 3.688 & 1.090 & 44.98 & 44.65 & 66.39 & \bf 69.36 &  1.536 \\
 & ShapeGF \cite{cai2020shapegf}
    & \bf 3.306 & \bf 1.027 & \bf 50.41 & \bf 47.12 & \bf 61.94 & \bf 70.51 &  \bf 1.059\\ 
 & \bf Ours     
    & \bf 3.276 & \bf 1.061 & \bf 48.71 & \bf 45.47 & \bf 64.83 & 75.12 & \bf 1.067 \\ 
\cmidrule{2-9}
 & Train
    & 3.917 & 1.003 & 51.73 & 54.04 & 48.85 & 50.82 & 0.809 \\

\midrule

\multirow{7}{*}{Chair}
 & PC-GAN \cite{achlioptas2018rgan}
    & 13.436 & 3.104 & 46.23 & 22.14 & 69.67 & 100.00 & 6.649 \\
 & GCN-GAN \cite{valsesia2018gcngan}
    & 15.354 & 2.213 & 39.84 & 35.09 & 77.86 & 95.80 & 21.708 \\
 & TreeGAN \cite{shu2019treegan} 
    & 14.936 & 3.613 & 38.02 &  6.77 & 74.92 & 100.00& 13.282  \\
 & PointFlow \cite{yang2019pointflow} 
    & 13.631 & 1.856 & 41.86 & 43.38 & 66.13 & \bf 68.40 & 12.474 \\ 
 & ShapeGF \cite{cai2020shapegf} 
    & \bf 13.175 & \bf 1.785 & \bf 48.53 & \bf 46.71 & \bf 56.17 & \bf 62.69 & \bf 5.996 \\ 
 & \bf Ours     
    & \bf 12.276 & \bf 1.784 & \bf 48.94 & \bf 47.52 & \bf 60.11 & 69.06 & \bf 7.797 \\ 
\cmidrule{2-9}
 & Train
    & 13.954 & 1.756 & 53.29 & 54.90 & 49.14 & 48.28 & 3.602 \\

\bottomrule
\end{tabular}
\end{center}
\end{table*}

\subsection{Point Cloud Auto-Encoder}
We implement a point cloud auto-encoder based on the probabilistic model in Section~\ref{sec:method}. 
We employ the PointNet as the representation encoder, denoted as $E_\vphi(\mX^\ts{0})$ with parameters $\vphi$, and leverage the reverse diffusion process presented in Section~\ref{sec:diffusion} for decoding, conditioned on the latent code produced by the encoder. 

Leveraging on Eq.~(\ref{eq:objective_final}), we train the auto-encoder by minimizing the following adapted objective:
\begin{equation}
\label{eq:objae}
\begin{split}
    L(\vtheta, \vphi) & = \E_q \Big[ \sum_{t=2}^{T} \sum_{i=1}^{N} \KL\big(q(\vx_i^\ts{t-1} | \vx_i^\ts{t}, \vx_i^\ts{0}) \| \\
    & \qquad \qquad \qquad \qquad p_\vtheta(\vx_i^\ts{t-1} | \vx_i^\ts{t}, E_\vphi(\mX^\ts{0})) \big) \\
    & \qquad
    - \sum_{i=1}^{N} \log p_\vtheta(\vx_i^\ts{0} | \vx_i^\ts{1}, E_\vphi(\mX^\ts{0}))
    \Big].
\end{split}
\end{equation}

To decode a point cloud that is encoded as the latent code $\vz$, we sample some points $\{ \vx_i^\ts{T}\}$ from the noise distribution $p(\vx_i^\ts{T})$ and pass the points through the reverse Markov chain $p_\vtheta(\vx_i^\ts{0:T} | \vz)$ defined in Eq.~(\ref{eq:reverse}) to acquire the reconstructed point cloud $\mX^\ts{0}=\{ \vx_i^\ts{0} \}_{i=1}^{N} $.

\begin{figure}[t]
\begin{center}
\includegraphics[width=0.45\textwidth]{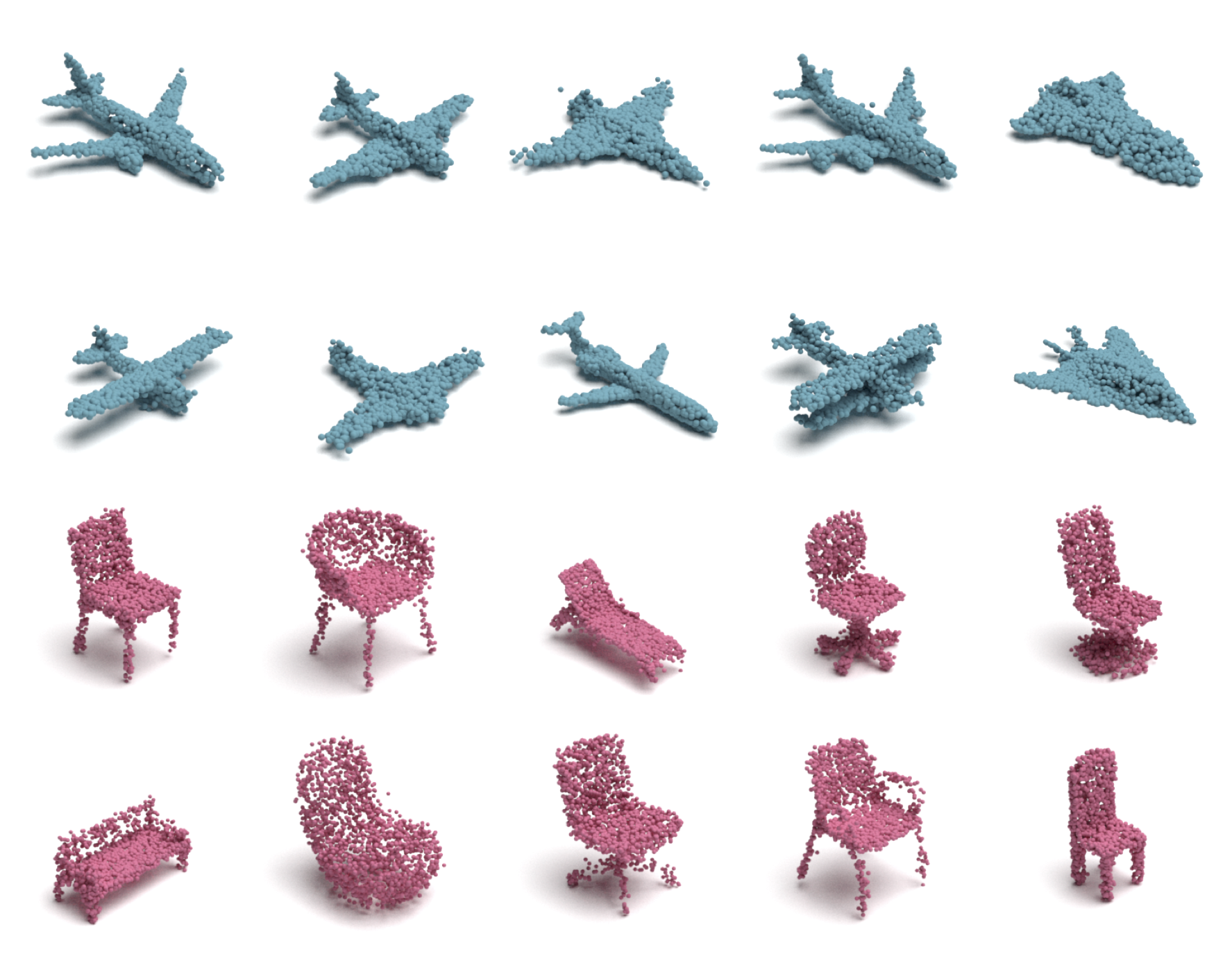}
\end{center}
\caption{Examples of point clouds generated by our model.}
\label{fig:visualization}
\end{figure}

\begin{table*}[t]
\caption{Comparison of point cloud auto-encoding performance. Atlas (S1) and Atlas (P25) denote 1-sphere and 25-square variants of AtlasNet respectively. CD is multiplied by $10^3$ and EMD is multiplied by $10^2$.}
\label{table:autoencode}
\begin{center}

\begin{tabular}{l c|cccc c|c}
\toprule
Dataset & Metric & Atlas (S1) & Altas (P25) & PointFlow & ShapeGF & Ours & Oracle \\ \midrule

\multirow{2}{*}{Airplane} & CD &
2.000 & \bf 1.795 & 2.420 & 2.102 & 2.118 & 1.016 \\
~ & EMD &
4.311 & 4.366 & 3.311 & 3.508 & \bf 2.876 & 2.141 \\
\midrule

\multirow{2}{*}{Car} & CD & 
6.906 & 6.503 & 5.828 & \bf 5.468 & 5.493 & 3.917 \\
 & EMD & 
5.617 & 5.408 & 4.390 & 4.489 & \bf 3.937 & 3.246 \\
\midrule

\multirow{2}{*}{Chair} & CD & 
5.479 & \bf 4.980 & 6.795 & 5.146 & 5.677 & 3.221 \\
 & EMD & 
5.550 & 5.282 & 5.008 & 4.784 & \bf 4.153 & 3.281 \\
\midrule

\multirow{2}{*}{ShapeNet} & CD & 
5.873 & 5.420 & 7.550 & 5.725 & \bf 5.252 & 3.074 \\
 & EMD & 
5.457 & 5.599 & 5.172 & 5.049 & \bf 3.783 & 3.112 \\
\bottomrule
\end{tabular}

\end{center}
\end{table*}

\begin{figure}[]
\begin{center}
\includegraphics[width=0.45\textwidth]{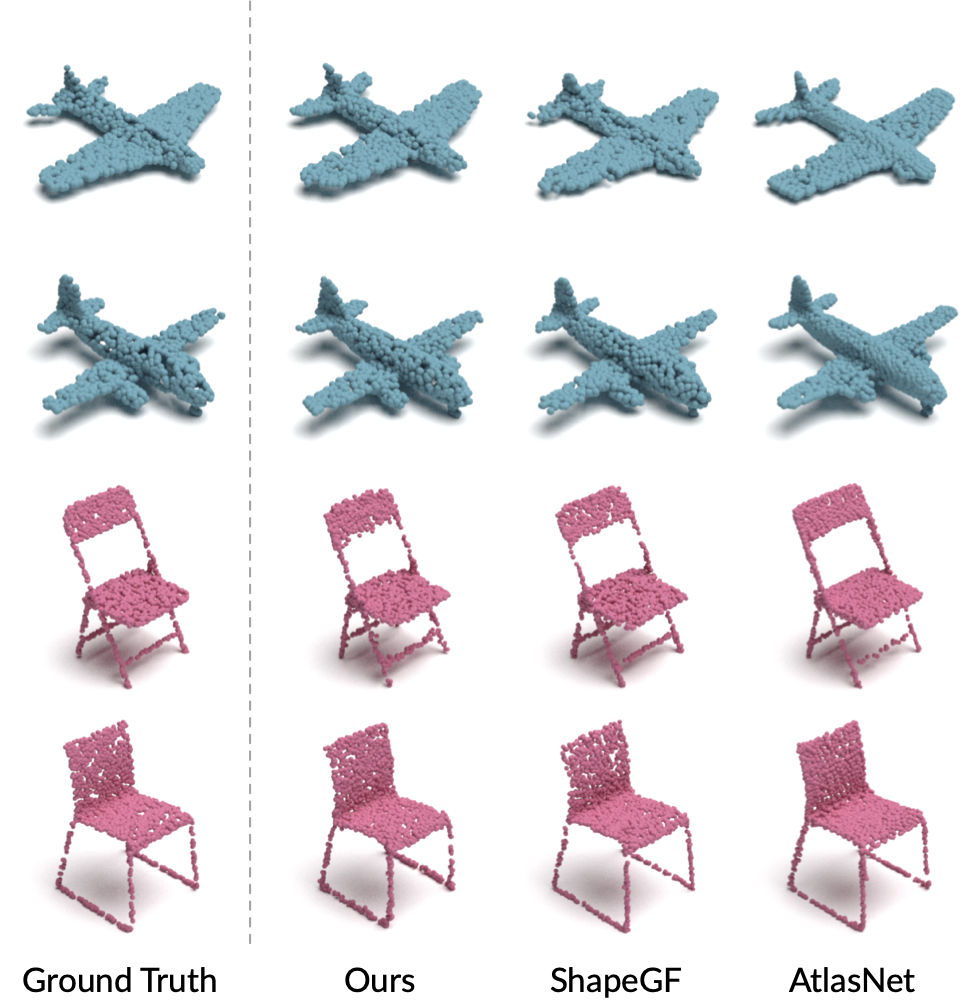}
\end{center}
\caption{Examples of reconstructed point clouds from different auto-encoders.}
\label{fig:autoencode}
\end{figure}

\section{Experiments}
In this section, we evaluate our model's performance on three tasks: point cloud generation, auto-encoding, and unsupervised representation learning. 

\subsection{Experimental Setup}
\label{sec:expoverview}

\paragraph{Datasets}
For generation and auto-encoding tasks, we employ the ShapeNet dataset \cite{shapenet2015} containing 51,127 shapes from 55 categories. 
The dataset is randomly split into training, testing and validation sets by the ratio 80\%, 15\%, 5\% respectively.
For the representation learning task, we use the training split of ShapeNet to train an encoder. Then we adopt ModelNet10 and ModelNet40 \cite{modelnet} to evaluate the representations learned by the encoder.
We sample 2048 points from each of the shape to acquire the point clouds and normalize each of them to zero mean and unit variance.

\paragraph{Evaluation Metrics}
Following prior works, we use the Chamfer Distance (CD) and the Earth Mover's Distance (EMD) to evaluate the reconstruction quality of the point clouds \cite{achlioptas2018rgan}. To evaluate the generation quality, we employ the minimum matching distance (MMD), the coverage score (COV), 1-NN classifier accuracy (1-NNA) and the Jenson-Shannon divergence (JSD) \cite{yang2019pointflow}. The MMD score measures the fidelity of the generated samples and the COV score detects mode-collapse. The 1-NNA score is computed by testing the generated samples and the reference samples by a 1-NN classifier. If the performance of the classifier is close to random guess, \ie, the accuracy is close to 50\%, the quality of generated samples can be considered better. The JSD score measures the similarity between the point distributions of the generated set and the reference set.

\begin{figure*}[]
\begin{center}
\includegraphics[width=0.9\textwidth]{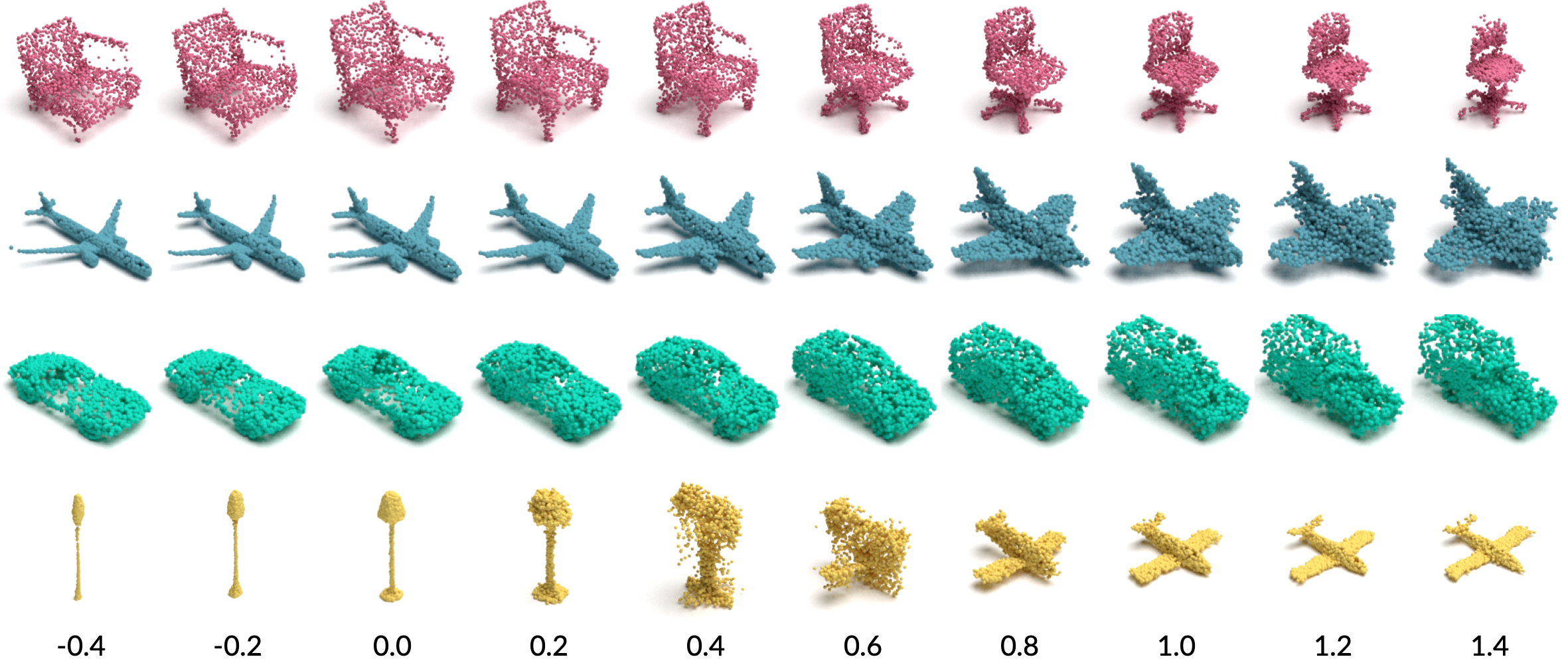}
\end{center}
\vspace{-0.1in}
\caption{Latent space interpolation and extrapolation.}
\label{fig:interp}
\end{figure*}
\vspace{-0.05in}

\subsection{Point Cloud Generation}

We quantitatively compare our method with the following state-of-the-art generative models: PC-GAN \cite{achlioptas2018rgan}, GCN-GAN \cite{valsesia2018gcngan}, TreeGAN \cite{shu2019treegan}, PointFlow \cite{yang2019pointflow} and ShapeGF \cite{cai2020shapegf} using point clouds from two categories in ShapeNet: \textit{airplane} and \textit{chair}. 
Following ShapeGF \cite{cai2020shapegf}, when evaluating each of the model, we normalize both generated point clouds and reference point clouds into a bounding box of $[-1, 1]^3$, so that the metrics focus on the shape of point clouds but not the scale.
We evaluate the point clouds generated by the models using the metrics in Section~\ref{sec:expoverview} and summarize the results in Table~\ref{table:generate}.
We also visualize some generated point cloud samples from our method in Figure~\ref{fig:visualization}.

\begin{table}[t]
\caption{Comparison of representation learning in linear SVM classification accuracy.}
\label{table:classify}
\begin{center}
\begin{tabular}{l|cc}
\toprule
\makecell[c]{Model} & ModelNet10 & ModelNet40 \\
\midrule
AtlasNet \cite{groueix2018atlasnet} 
    & 91.9 & 86.6 \\
PC-GAN (CD) \cite{achlioptas2018rgan}
    & \bf 95.4 & 84.5 \\
PC-GAN (EMD) \cite{achlioptas2018rgan}
    & \bf 95.4 & 84.0 \\
PointFlow \cite{yang2019pointflow}
    & 93.7 & 86.8 \\
ShapeGF \cite{cai2020shapegf}
    & 90.2 & 84.6 \\
\midrule
Ours
    & 94.2 & \bf 87.6 \\
\bottomrule
\end{tabular}

\end{center}
\end{table}

\begin{figure}[]
\begin{center}
\includegraphics[width=0.45\textwidth]{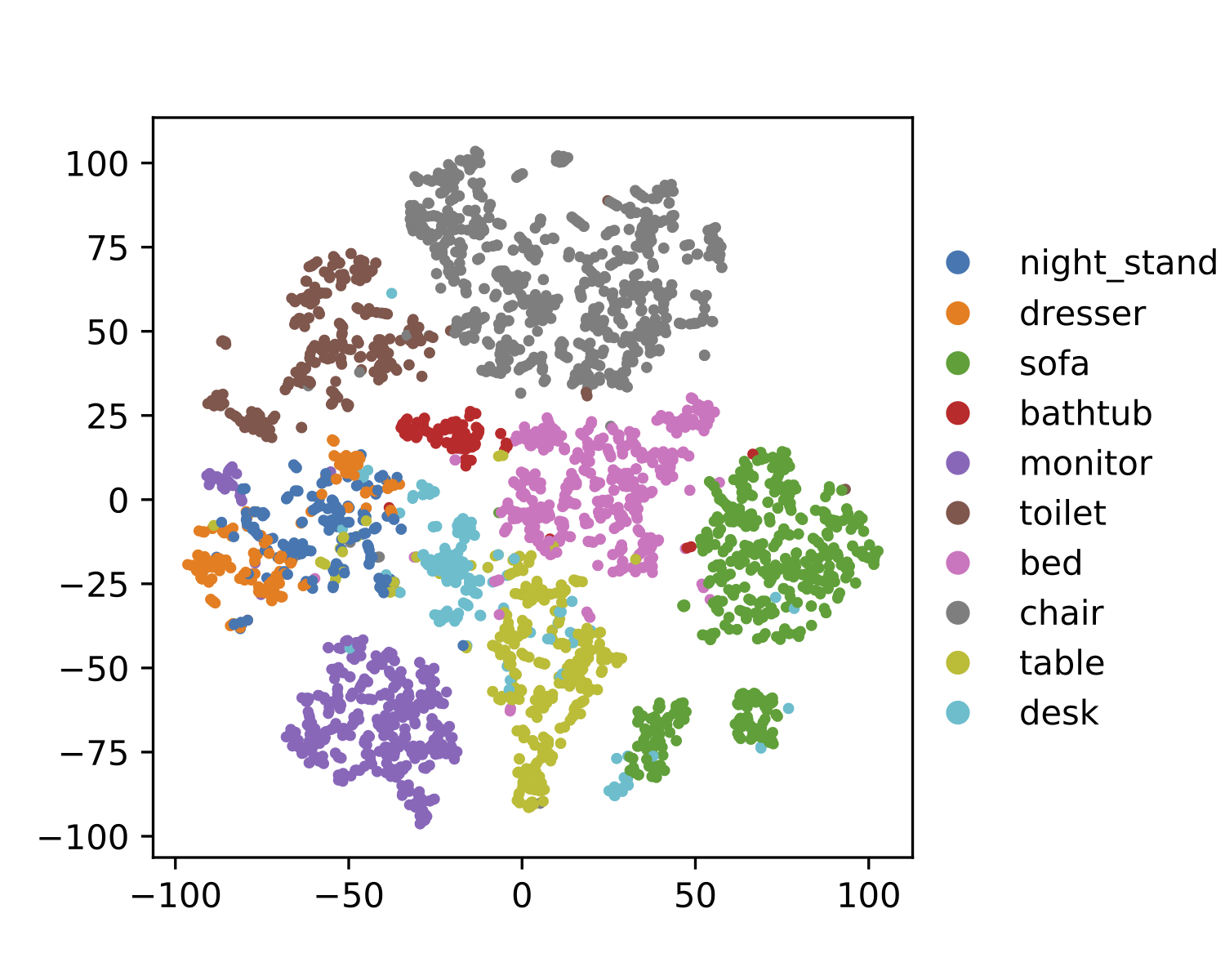}
\end{center}

\caption{The t-SNE clustering visualization of latent codes obtained from the encoder.}
\label{fig:tsne}
\end{figure}

\subsection{Point Cloud Auto-Encoding}

We evaluate the reconstruction quality of the proposed auto-encoder, with comparisons against state-of-the-art point cloud auto-encoders: AtlasNet \cite{groueix2018atlasnet}, PointFlow \cite{yang2019pointflow} and ShapeGF \cite{cai2020shapegf}. 
Four datasets are used in the evaluation, which include three categories in ShapeNet: \textit{airplane}, \textit{car}, \textit{chair} and the whole ShapeNet. 
We also report the lower bound ``oracle'' of the reconstruction errors. This bound is obtained by computing the distance between two different point clouds with the same number of points and the identical shape.
As shown in Table~\ref{table:autoencode}, our method outperforms other methods when measured by EMD, and pushes closer towards the lower bounded ``oracle'' performance. 
The CD score of our method is comparable to others. Notably, when trained and tested on the whole ShapeNet dataset, our model outperforms others in both CD and EMD, which suggests that our model has higher capacity to encode different shapes. 
Also, the visualization of reconstructed point clouds in Figure~\ref{fig:autoencode} validates the effectiveness of our model.

\subsection{Unsupervised Representation Learning}
\vspace{-0.05in}

Further, we evaluate the representation learned by our auto-encoder. 
Firstly, we train an auto-encoder with the whole ShapeNet dataset. During the training, we augment point clouds by applying random rotations along the gravity axis, which follows previous works. 
Then, we learn the feature representations of point clouds in ModelNet-10 and ModelNet-40 using the trained encoder, and train a linear SVM using the codes of point clouds in the training split and their categories. 
Finally, we test the SVM using the testing split and report the accuracy in Table~\ref{table:classify}. 
We run the official code of AtlasNet and ShapeGF to obtain the results, since the results are not provided in their papers. For PC-GAN and PointFlow, we use the results reported in the papers.
The performance of our encoder is comparable to related state-of-the-art generative models.

In addition, we project the latent codes of ModelNet-10 point clouds produced by the encoder into the 2D plane using t-SNE \cite{maaten2008tsne}, and present it in Figure~\ref{fig:tsne}. 
It can be observed that there are significant margins between most categories, which indicates that our model manages to learn informative representations. 
Further, we visualize the interpolation and extrapolation between latent codes in Figure~\ref{fig:interp}.

\section{Conclusions}

We propose a novel probabilistic generative model for point clouds, taking inspiration from the diffusion process in non-equilibrium thermodynamics. 
We model the reverse diffusion process for point cloud generation as a Markov chain conditioned on certain shape latent, and derive a tractable training objective from the variational bound of the likelihood of point clouds.   
Experimental results demonstrate that the proposed model achieves the state-of-the-art performance in point cloud generation and auto-encoding.


{\small
\bibliographystyle{ieee_fullname.bst}
\bibliography{references}
}

\end{document}